\documentclass[letterpaper, 10 pt, conference]{ieeeconf}  

\IEEEoverridecommandlockouts                              

\usepackage{mathptmx} 
\usepackage{times} 
\usepackage{amsmath} 
\usepackage{amssymb}  
\usepackage{balance}
\usepackage{graphicx}
\usepackage{nth}
\usepackage[hidelinks]{hyperref}
\usepackage{siunitx} 
\usepackage{bm} 
\usepackage{subcaption} 
\usepackage{algorithm2e} 
\usepackage{xcolor} 
\usepackage{tabularx} 
\usepackage{dcolumn} 

\title{\LARGE \bf
Design, Simulation, and Testing of a Flexible Actuated Spine for Quadruped Robots
}

\author{Andrew P. Sabelhaus$^{1}$*, {\it{Student Member, IEEE}}, Lara Janse van Vuuren$^{1}$, Ankita Joshi$^{1}$, \\ Edward Zhu$^{1}$, Hunter J. Garnier$^{1}$, Kimberly A. Sover$^{1}$, Jesus Navarro$^{1}$, \\ Adrian K. Agogino$^{2}$, Alice M. Agogino$^{1}$, {\it{Senior Member, IEEE}}
\thanks{*Authors with the NASA Ames Research Center Intelligent Robotics Group, 
        Moffett Field, CA 94035, USA}
\thanks{$^{1}$A.P. Sabelhaus, L. Janse van Vuuren, A. Joshi, E. Zhu, H.J. Garnier, K.A. Sover, J. Navarro, and A.M. Agogino are with the Department of Mechanical Engineering,
        University of California Berkeley, USA
        {\tt\small apsabelhaus, larajvv, ankitaj, edward.zhu, hgarnier, kasover, foamjay, agogino @berkeley.edu}}%
\thanks{$^{2}$A.K. Agogino is with the Intelligent Systems Divison,
        NASA Ames Research Center, Moffet Field CA 94035
        {\tt\small adrian.k.agogino@nasa.gov }}%
}

\begin{document}

\maketitle
\thispagestyle{empty}
\pagestyle{empty}

\begin{abstract}



Walking quadruped robots face challenges in positioning their feet and lifting their legs during gait cycles over uneven terrain.
The robot Laika is under development as a quadruped with a flexible, actuated spine designed to assist with foot movement and balance during these gaits.
This paper presents the first set of hardware designs for the spine of Laika, a physical prototype of those designs, and tests in both hardware and simulations that show the prototype's capabilities.
Laika's spine is a tensegrity structure, used for its advantages with weight and force distribution, and represents the first working prototype of a tensegrity spine for a quadruped robot.
The spine bends by adjusting the lengths of the cables that separate its vertebrae, and twists using an actuated rotating vertebra at its center.
The current prototype of Laika has stiff legs attached to the spine, and is used as a test setup for evaluation of the spine itself.
This work shows the advantages of Laika's spine by demonstrating the spine lifting each of the robot's four feet, both as a form of balancing and as a precursor for a walking gait.
These foot motions, using specific combinations of bending and rotation movements of the spine, are measured in both simulation and hardware experiments.
Hardware data are used to calibrate the simulations, such that the simulations can be used for control of balancing or gait cycles in the future. 
Future work will attach actuated legs to Laika's spine, and examine balancing and gait cycles when combined with leg movements.

\end{abstract}

\section{INTRODUCTION}

\begin{figure}[thpb]
    \centering
    \includegraphics[width=1\columnwidth]{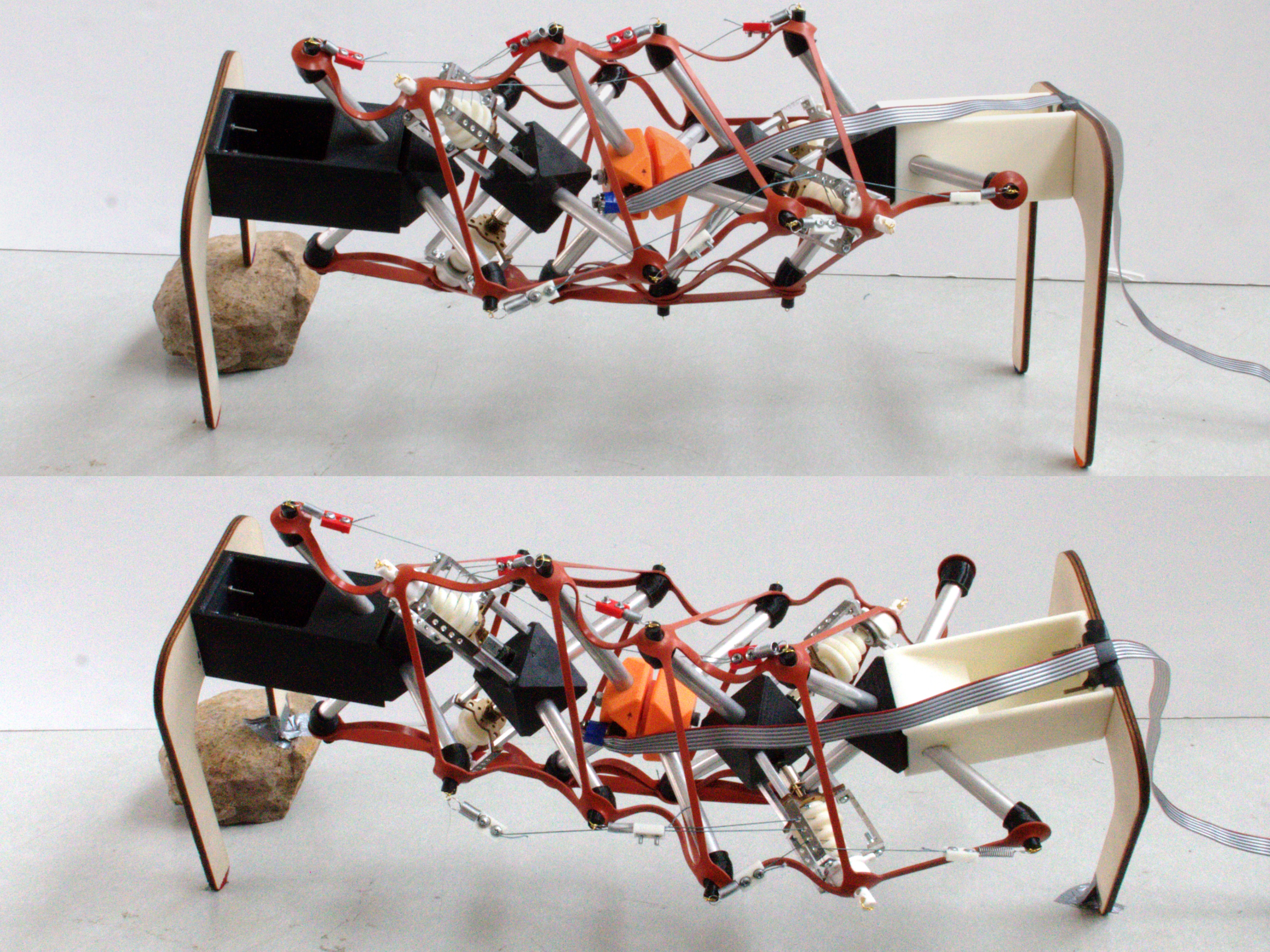}
    \caption{Prototype of the Laika quadruped robot with the flexible, actuated tensegrity spine presented in this paper. Motion of the spine allows a robot to balance on obstacles such as rocks (top). In comparison, without actuating the spine, the robot cannot balance and falls to its side (bottom.) This test is performed using an obstacle that is almost half the hip-height of the robot (7.5 cm vs. 18.5 cm.)}
    \label{fig:overview}
    \vspace{-0.4cm}
\end{figure}


Walking over uneven terrain is a common goal of many robotic systems, especially for quadruped (four-legged) robots \cite{Kimura1999,Fukuoka2003,Kolter2007,Palmer2007,Pongas2007,Raibert2008,Buchli2009,Hirose2009}. 
Such terrain could be as diverse as the stairs inside buildings \cite{Hirose2009,Sprowitz2013,Ev2016} or rocky planets.
However, balancing and locomotion over large obstacles can be challenging for robots built with rigid structures that cannot conform to the environment, limiting them to obstacles that are small in comparison to their total size \cite{Gehring2014,Fukuoka2003}.

This research presents a flexible, actuated spine for quadruped robots that can assist in a robot's locomotion over uneven terrain in space applications.
The spine is used here as part of an early prototype of the robot Laika, named after the first dog in space, as adapted from earlier concepts \cite{Sabelhaus2015}.

This work presents the first design of a `tensegrity' spine in a quadruped robot, the first hardware prototype of that design, and the first experimental results (from both simulation and hardware) which show the potential advantages of the spine.
These advantages are shown in the form of lifting each of the robot's four feet, using specific combinations of bending and rotation of the spine.
Lifting the feet of the robot allows it to balance on obstacles (Fig. \ref{fig:overview}).
This is also the first movement in a quadruped gait cycle \cite{Furusho1995,Fukuoka2003}.
Actuated legs will be added for walking locomotion in future work.



\section{Prior Work}

Laika's spine is a tensegrity structure, with five independent vertebrae held together in a tension network.
Tensegrity (or ``tension-integrity") systems have rigid elements, in compression, suspended in a network of cables in tension \cite{Paul2006a}.
The spine changes its shape by adjusting the lengths of the cables that hold its vertebrae apart, as do other tensegrity robots \cite{Sabelhaus2015a,friesen2014,Caluwaerts2015,Friesen2016}.
This model of the spine has actuators that adjust the lengths of multiple sets of cables simultaneously, creating pre-defined bending motions.


A large body of other work on quadruped robots has incorporated a spine in different ways. \cite{Takuma2010,Kani2011,Khoramshahi2013,Park2014a,Horvat2015,Eckert2015,Sprowitz2013,Ugurlu2013a}, including passive and actuated designs for different purposes.
Laika and its spine differ from prior quadrupeds with spines in multiple important ways.

First, Laika's spine is both actively actuated and passively compliant.
Other passive spines that assist with locomotion cannot actively move to balance the robot \cite{Takuma2010,Kani2011,Seok2014}.
In contrast, robots with actuated spines often do not include compliance or flexible mechanisms \cite{Berns1998, Berns1999}, enforcing significant design constraints for walking over obstacles.

Second, Laika's spine can perform complex and multi-degree-of-freedom motions simultaneously, including the main three studied in spine robots: axial rotation, and bending in both the sagittal and coronal planes.
Other robots with both flexible and fixed-twisting spines are often designed for to sagittal-only \cite{Khoramshahi2013,Eckert2015} or coronal-only \cite{Weinmeister2015} bending.


Third, Laika's spine is designed for balancing and adjusting the robot's foot placement.
Other spines are more commonly used as a part of a dynamic running or galloping gait  \cite{Khoramshahi2013,Park2014a,Eckert2015}, or as an input to change the direction of running \cite{Weinmeister2015}.

Finally, this is the first spine built as a tensegrity system.
This approach includes adjustable compliance, passive force distribution, redundancy and robustness to failure, and lightweight construction \cite{Caluwaerts2014,Sabelhaus2015,Sabelhaus2015a}.

\section{ROBOT STRUCTURE}

The current design of Laika consists of a spine supported by rigid hips, shoulders, and legs.
As its spine moves, the center of mass shifts in three dimensions, enabling lifting of each foot accordingly.
The design of the robot's vertebrae, and its tendon placement, evolved from both biological concepts as well as iteration on designs that could create all  three basic motions \cite{Sabelhaus2015}. 
Actuated legs will be added in future work to test the spine with locomotion.


\subsection{Spine Topology}\label{sec:spine_topology}

\begin{figure}[thpb]
    \centering
    \includegraphics[width=1\columnwidth]{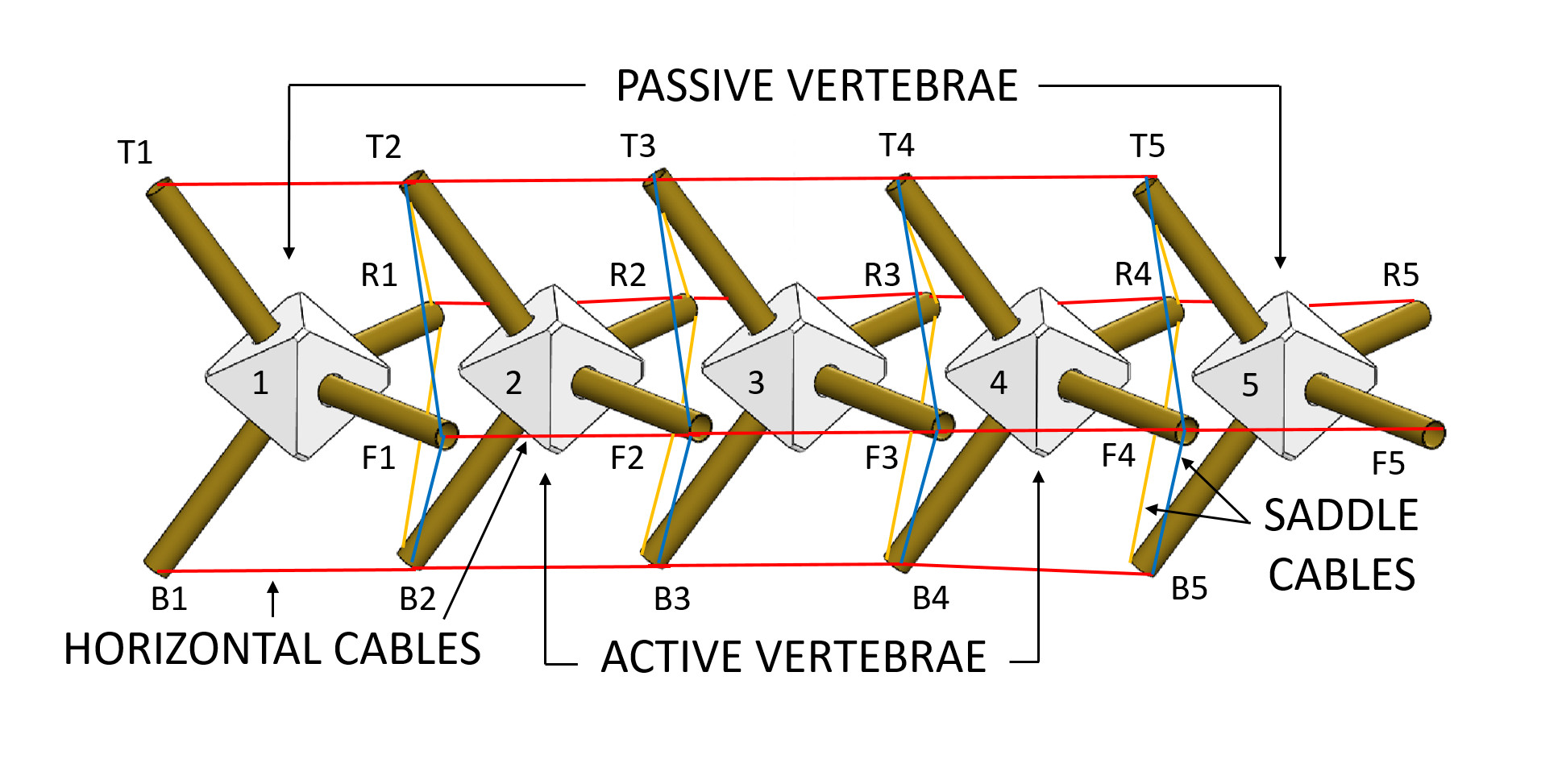}
    \caption{Spine Topology. Laika's spine consists of five tetrahedral vertebrae with two types of connecting cables (\textit{horizontal} and \textit{saddle}).}
    \label{fig:spine_topology}
    \vspace{-0.4cm}
\end{figure}

Laika's spine is composed of five vertebrae, out of which the second and the fourth vertebrae are \textit{active}, with actuators, and the third vertebra in the center provides rotation. 
Vertebrae 1 and 5 connect to the hips and shoulders, and are \textit{passive}, without actuation.
The vertebrae are connected by a lattice of cables in tension which balance and stabilize the structure. 
The active vertebrae contain motors which adjust the lengths of these cables. 
Fig. \ref{fig:spine_topology} shows the lattice network of cables for the spine, where the attachment points for the cables are labeled based on the vertebra and the side of the spine they belong to (for example, the attachment point on the front side of vertebra 1 is labeled F1).

The horizontal cables connect the different vertebrae on each of the four sides of the spine: top, bottom, left, or right, respectively (Fig. \ref{fig:spine_topology}). Each set of horizontal cables consists of four individual cables, running from one vertebra's motor to attachment points on each of the other four vertebrae. For example, in Fig. \ref{fig:spine_topology}, four separate cables run from the actuated end at T2 to T1, T3, T4, and T5. 


Four \textit{saddle} cables connect one end of a vertebra to two different ends of the adjacent vertebra, and are required to provide opposing tension forces from the horizontal cables.
For example, the F1 end from vertebra one is connected to ends T2 and B2 on vertebra two (Fig. \ref{fig:spine_topology}). 

Like other tensegrity structures, the spine changes its shape by adjusting the lengths of its cables. 
From this topology, it can be seen that shortening one set of horizontal cables causes the vertebrae to move closer together along one edge, resulting in a bending motion in either of the robot's sagittal or coronal planes.



\subsection{Hardware Overview}\label{sec:hardware_platform}

\begin{figure}[hb]
    \centering
    \includegraphics[width=1\columnwidth]{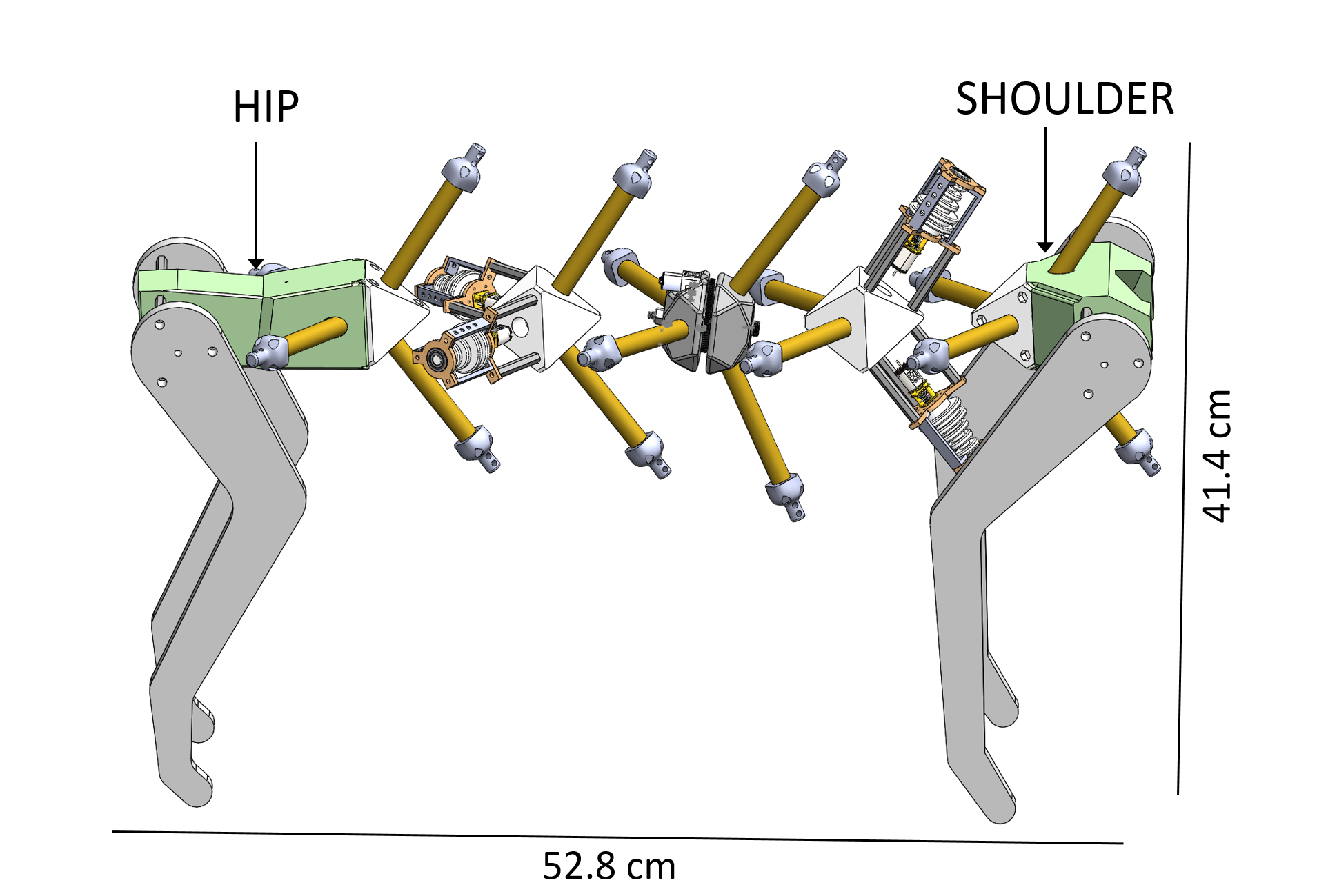}
    \caption{Robot Assembly. The spine is mounted onto the four legs attached to the hip and shoulder. This CAD render does not show the cables.}
    \label{fig:assembly}
\end{figure}

The current prototpe of Laika consists of the spine connected to a 3D printed hip and shoulder that stand on stiff, rapid-prototyped legs.
Varying sizes of legs can be attached.
A representative model of Laika (Fig.  \ref{fig:assembly}) is approximately 52.8 cm long and stands 41.4 cm tall, and weighs 1.62 kg. 
A more compact version is used for testing of the spine.

The robot's tension-network cables are implemented using a combination of two structural elements.
First, a passive under-structure, built from a laser-cut elastomer lattice (as described in \cite{Chen2017}) keeps the robot in even tension.
Then, a set of stiff cables, attached to mechanical springs, are used for actuation.
The horizontal cables are actuated, whereas the saddle cables are only held in place by the elastomer lattice.

Two different materials are used for the lattice along different edges.
The majority of the robot's lattice is a silicone rubber, seen as the orange material in Figs. \ref{fig:overview} and \ref{fig:spool_hardware}.
An additional Buna-N rubber, with a higher stiffness, is used for the ventral horizontal cable.
This stiffer strip of lattice counteracts the robot's weight.

There are thus three sets of properties for modeling the robot's cables.
These are summarized in table \ref{tab:material_props}.
The elastomers are inexpensive materials for which no manufacturer data was available; thus, a set of tests were done to estimate a linear spring constant.
The variability in these constants serves as a way to calibrate the simulations in sec. \ref{sec:results}.


\begin{table}[!hb]
\centering
\caption{Hardware Prototype Material Properties.}
\label{tab:material_props}
\begin{tabularx}{1\columnwidth}{| l | l | X |}
    \hline
    Cable Material & Spring Constant ($N/m$) & Std Dev. ($N/m$) \\
    \hline
    Silicone Rubber & 237 & 11 \\
    Buna-N Rubber & 810 & 132 \\
    Mechanical Springs & 187 & -- (exact) \\
    \hline
\end{tabularx}
\end{table}

All actuated components of this prototype use a brushed DC motor with a 1000:1 gearbox, are position-controlled using an encoder.


\subsection{Vertebra and Actuator Design}\label{sec:vertebra_design}

\begin{figure}[h]
    \centering
    \vspace{-0.1cm}
    \includegraphics[width=0.7\columnwidth]{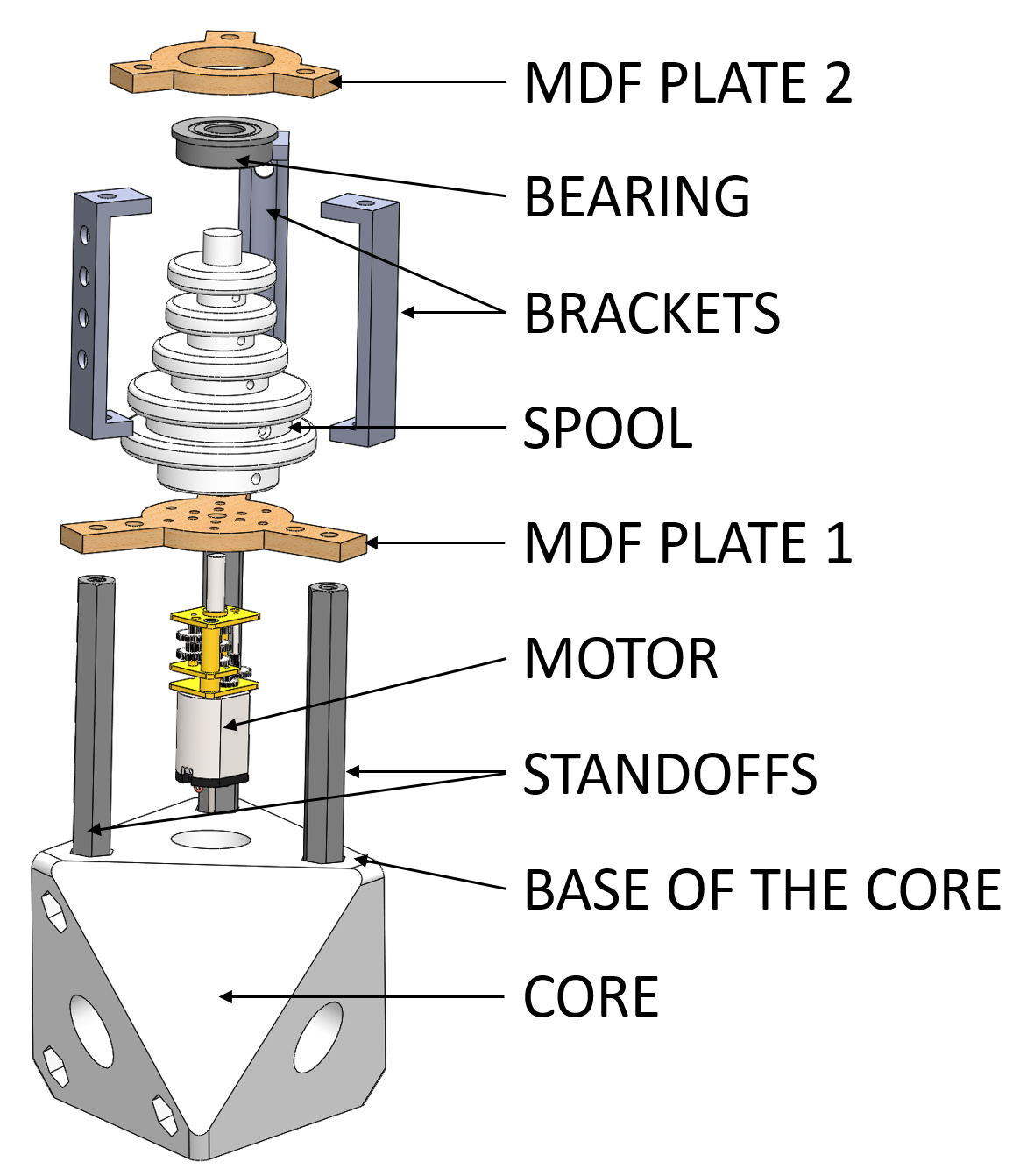}
    \caption{Actuator assembly. Each active vertebrae has two of these actuators that connect a motor to a spool in order to adjust the lengths of the horizontal cables.}
    \label{fig:spool_CAD}
    \vspace{-0.1cm}
\end{figure}

\begin{figure}[h]
    \centering
    \includegraphics[width=0.9\columnwidth]{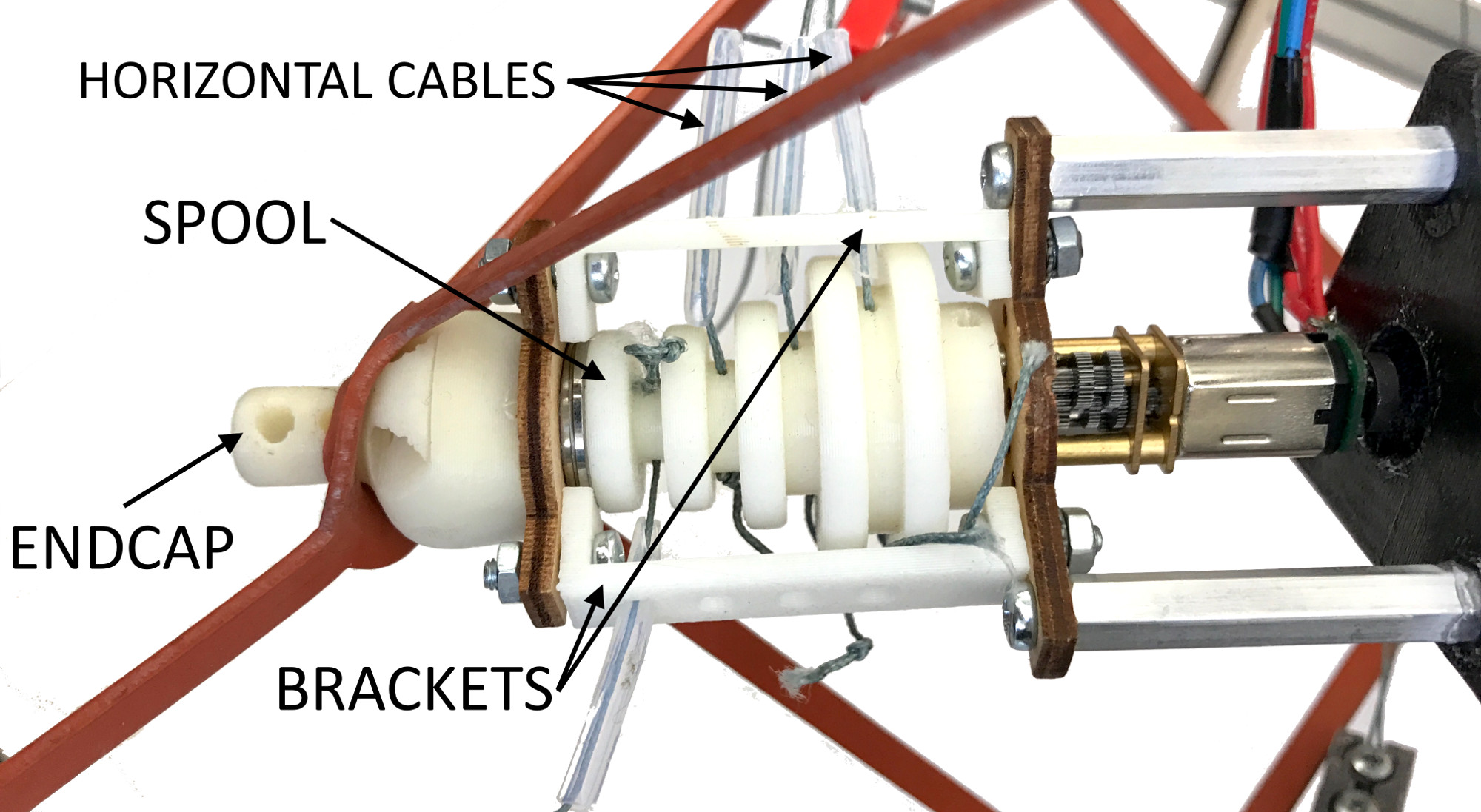}
    \caption{Spool prototype on an active vertebra. The spool is 3D printed such that the diameters of the gears match the varying horizontal cables lengths, as originally calculated using inverse kinematics in \cite{Sabelhaus2015}.} 
    \label{fig:spool_hardware}
    \vspace{-0.2cm}
\end{figure}

Each vertebra has a 3D printed core which holds either the actuator assemblies (active vertebrae) or lattice and spring attachment points (on both passive and active.) 
For the passive vertebrae, the core has four rods with end caps that hold the attachment points.
The active vertebrae have two rods with end caps, and two motor assemblies. 
These actuator assemblies have a bracket that connects to the core, and holds the motor and cable spool (Fig. \ref{fig:spool_CAD}). 
A hardware prototype of one actuator inside the elastomer lattice is shown in Fig.  \ref{fig:spool_hardware}.
CAD files are provided online which  include specific dimensions for each component\footnote{https://github.com/BerkeleyExpertSystemTechnologiesLab/ultra-spine-hardware}.

The cable actuators for Laika's spine are designed to create a motion primitive (bending) with only one input.
In prior work, simulations using inverse kinematics of the spine showed that this can be achieved by adjusting the cable lengths at a fixed ratio with respect to each other \cite{Sabelhaus2015}.
This work implements that design using a 3D printed spool which has four different grooves, one for each cable in a given set of four horizontal cables (Fig. \ref{fig:spool_hardware}). 

The diameters of these spool grooves correspond to the length-change ratio from \cite{Sabelhaus2015} for a constant-radius bend, a ratio of 1-1-2-3.
When the spool rotates, each cable's length changes proportionally to the diameter of that groove, retracting the cables at different rates. 

\subsection{Rotating Vertebra Design}\label{sec:rot_vert_design}

\begin{figure}[htb]
    \centering
    \includegraphics[width=0.9\columnwidth]{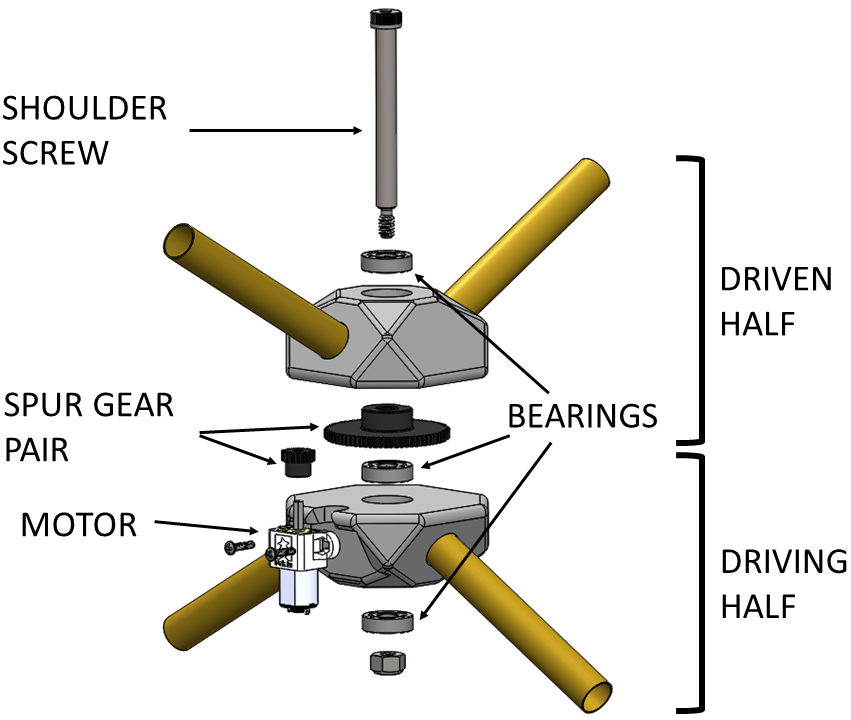}
    \caption{Rotating vertebra structure, used as vertebra 3 in the prototype. The moving halves of the structure allow for axial rotation of the spine.}
    \label{fig:rotating_core}
    \vspace{-0.2cm}
\end{figure}

An additional, different actuator is included as the middle vertebra of the spine, and provides its rotational degree-of-freedom (Fig. \ref{fig:rotating_core}.)
This vertebra is composed of two halves, one driving and one driven, which are connected through a shoulder screw that also acts as the shaft. 
The same motor as in Fig. \ref{fig:spool_CAD} and \ref{fig:spool_hardware} is mounted on the driving half and its torque is transmitted through a 4:1 spur gear pair to the driven half. 
This design was chosen for its structural and actuation simplicity, allowing for the straightforward generation of rotational motion.

\section{SIMULATION SETUP}


In order to use Laika's spine to balance and lift its feet, in anticipation of developing a walking gait, a simulation was developed that can predict foot position as a function of spine motions.
Although there is some intuition about which motions of the spine would lift which foot (for example, how the center of mass shifts as the robot bends), these simulations were used to quantify exactly how much movement would be required for different foot positions.
These tests were performed using the NASA Tensegrity Robotics Toolkit\footnote{http://irg.arc.nasa.gov/tensegrity/ntrt}.

Though both simulation and hardware have the capability to bend the robot in the sagittal plane, pulling the robot's dorsal (top) and ventral (bottom) cables, only coronal (left/right) bending is shown in this proof-of-concept.
Shifts in the spine's center-of-mass in the coronal plane are what differentiate which foot is lifted.


\subsection{Simulation Environment}

The NASA Tensegrity Robotics Toolkit (NTRT, or NTRTsim) is an open source package for modeling, simulation, and control of tensegrity robots based on the Bullet Physics engine \cite{BulletPhysicsEngine2013}.
Prior work has validated both the kinematics \cite{Caluwaerts2014} and dynamics \cite{Mirletz2015} of the simulator, and it has been extensively used in prior tensegrity robotics work \cite{friesen2014,Friesen2016,Lessard2016,Kim2014,Chen2016,Sabelhaus2015,Mirletz2015a}.

Cables tensions are modeled in NTRT as virtual spring-dampers, as in:

\vspace{-0.2cm}

\begin{equation}\label{eq:cable_model}
T_i = k (x_i - r_{i}) - c \dot x_i
\end{equation}

\noindent where $T_i$ is the tension force applied by cable $i$ when its total length is $x_i$, and the spring's rest length is $r_{i}$.
The simulations controlled the rest lengths $r_i$, as if motors retracted the cables.




\subsection{Robot Model in NTRT}

\begin{figure}[thpb]
    \centering
    \includegraphics[width=1.0\columnwidth]{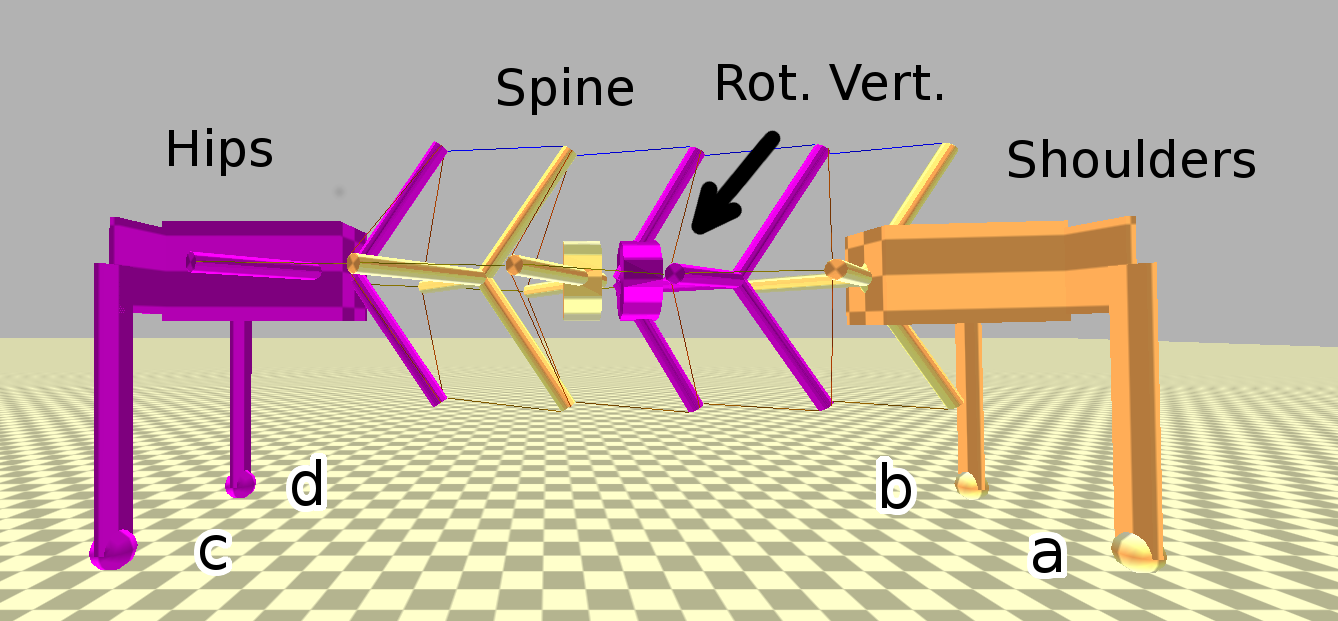}
    \caption{Model of Laika and its spine in the NASA Tensegrity Robotics Toolkit simulation, with the rotating vertebra. Feet labeling are: (a) front right, (b) front left, (c) back right, (d) back left.}
    \label{fig:simulation_overview}
    \vspace{-0.4cm}
\end{figure}

Like the hardware, the robot model consists of a spine with its rotating vertebra plus shoulders, hips, and legs (Fig. \ref{fig:simulation_overview}). 
The spine is rendered as a simplified model consisting of cylindrical rods.
This is similar to the structure used in \cite{Sabelhaus2015,Mirletz2015a}.
The rotating vertebra is constructed out of two constrained rigid bodies, placed inside the spine, and is position-controlled by specifying a rotation between the two halves.

\subsection{Test Procedure}\label{sec:test_trajectories}

The initial simulations presented in this work represent a small subset of the robot's motions that may lift a foot, chosen in order to make a hardware comparison tractable.
Specifically, bending in the coronal plane is restricted to two simple motions (left, or right), leaving the rotational degree-of-freedom as the primary variable.

Thus, simulation tests followed a two-step procedure.
First, one set of horizontal cables were retracted to a percent $P$ of their original rest length, as in:


\begin{equation}
r_i(t) = P \; r_i(0)
\end{equation}

\noindent in order to create a set left-or-right bend. Using observations from a first round of simulations, and the prototype itself, a horizontal length-change of $P = 80\%$ best represented a bending motion that sufficiently shifted the spine's center-of-mass.

Then, once the robot settled, the center vertebra was slowly rotated until one foot left the ground.
Positions of the feet were recorded alongside the angle of rotation of the center vertebra.
The center vertebra rotations for quasi-static motion were commanded as a slow ramp-input over $40$ sec. up to $60$ deg. in either direction, as in:

\begin{equation}
\theta(t) = \pm \left( \frac{t}{40} \right) \left( \frac{\pi}{3} \right)
\end{equation}

\subsection{Simulation Test Points for Calibration}


Multiple tests were performed in order to calibrate the simulation parameters against hardware, as is common for this simulator \cite{Caluwaerts2014}.
Since one of the major assumptions in the simulation is the linearity of the cable spring force, the variation in the spring constant for the robot's cables (Table \ref{tab:material_props}) provides a convenient method of calibration.

Five spring constants were tested with each of the four foot-lifting motions, varying both the silicone and Buna-N cable constants simultaneously.
This adjusts the spine's overall tension, calibrating for both the unknowns and nonlinearities in the materials as well as modeling simplifications.
Table \ref{tab:simulation_tests} shows the spring constants chosen, evenly spaced from -2 to +2 standard deviations of the mean from Table \ref{tab:material_props}.

\begin{table}[th]
\centering
\caption{Five spring constant test points (in $N/m$) for the simulation, adjusting the overall tension of the robot.}
\label{tab:simulation_tests}
\begin{tabularx}{1\columnwidth}{| l | X  X  X  X X |}
    \hline
    Material & Low ($-2\sigma$) & Med-Low & Mean ($\mu$) & Med-High & High ($+2\sigma$)\\
    \hline
    Silicone & 216 & 227 & 237 & 248 & 258 \\
    Buna-N & 547 & 678 & 810 & 941 & 1073 \\
    \hline
\end{tabularx}
\end{table}

\section{HARDWARE TEST SETUP}




Hardware tests were performed to show Laika's spine lifting its feet and to calibrate the simulation for future use.
Multiple tests were performed per foot, all using the same lattice under the same conditions.


\subsection{Hardware Testing Platform}

\begin{figure}[bhpt]
    \centering
    \includegraphics[width=1.0\columnwidth]{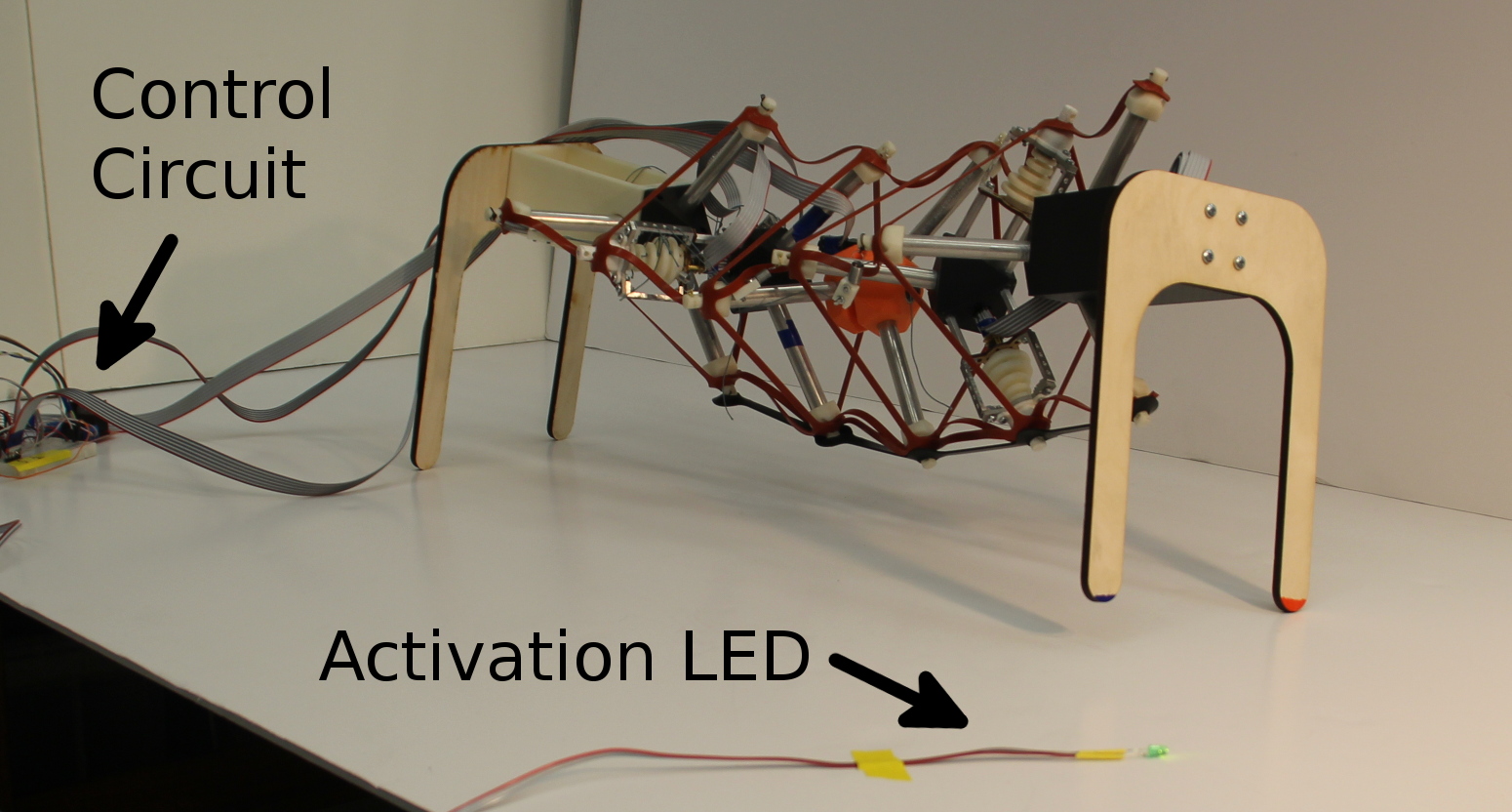}
    \caption{One test of Laika's spine. An offboard control circuit powered the motors and tracked rotations, and an LED was activated at the start of the test such that the video camera (out-of-frame) could correlate timestamps.}
    \label{fig:hardware_test_setup}
\end{figure}

The prototype of Laika was set up in placed of a camera, with off-board power and control (Fig. \ref{fig:hardware_test_setup}).
The robot's control system consisted of microcontroller connected to a power supply, with two connected motors: one for a horizontal cable set, and one for the center vertebra.
Power cables were wound through the spine to connect the motors to the controller.
During testing, the motors' encoders were used to track rotations, which were converted to percent-length-change in the controller.

\subsection{Hardware Test Procedure}

For each test, the spine was actuated along the same trajectories as in sec. \ref{sec:test_trajectories} for the simulation.
Switching between tests involved rotating the robot, re-routing the motors' cables, and if needed, attaching and detaching the required horizontal cables.
The robot was rotated by 180 degrees when switching between the tests for its anterior and posterior ends.
This required re-routing the motors' cables, causing a slight change in center-of-mass.

Prior to each test, the single set of actuated horizontal cables were tensioned until just past slackness, as an approximation to the simulation's initial conditions. 
Markers were placed on the bed of the test setup, and the robot was re-positioned to the same state between tests.






For each test, the video camera recorded the feet as the controller tracked the rotations.
A small LED, placed within the viewing frame of the camera, was activated at the start of the test, and a time series of data was recorded from the microcontroller.
After each test, the video was analyzed for the time at which the LED activated, and at which the desired foot just began to lift.
This time difference was then indexed into the controller data to find the rotation at that time.

\vspace{-0.2cm}
\section{RESULTS}\label{sec:results}

\begin{figure*}[ht!]
    \centering
    \includegraphics[width=0.8\textwidth]{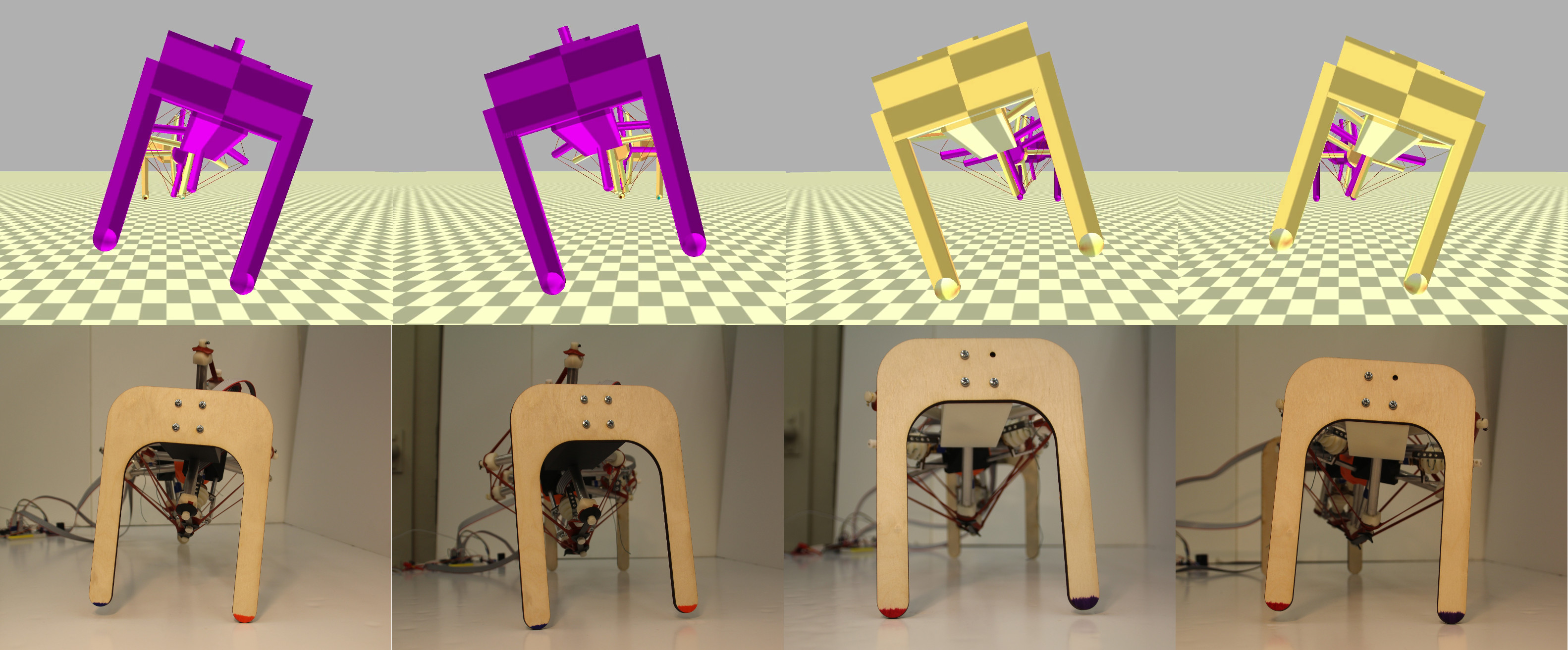}
    \caption{Lifting of each of Laika's feet in simulation (top) and hardware (bottom). Images left-to-right are for feet A, B, C, D (Front Right, Front Left, Back Right, Back Left). Hardware images taken just as liftoff occurs.}
    \label{fig:foot_lifting_vis}
    \vspace{-0.3cm}
\end{figure*}

By choosing four combinations of rotation direction and coronal-plane bending, Laika's spine was able to lift each of its four feet.
Each of the four motions are summarized in Table \ref{tab:motions}, and shown in Fig.  \ref{fig:foot_lifting_vis} on the following page.
These motions can be interpreted as a rotation lifting one diagonal set of legs, and bending shifting the robot's mass to raise one foot or the other.


\begin{table}[!hb]
\centering
\caption{Motion combinations of spine for foot lifting.}
\label{tab:motions}
\begin{tabularx}{1\columnwidth}{| l | l | X |}
    \hline
    Bend Dir. / Cabled Pulled &  Vert. Rotation Dir. & Foot Lifted? \\
    \hline
    Left Bend / Horiz. Right & (+), CCW & A, Front Right \\
    Right Bend / Horiz. Left & (+), CCW & C, Back Left \\
    Left Bend / Horiz. Right & (-), CW & B, Front Left \\
    Right Bend / Horiz. Left & (-), CW & D, Back Right \\
    \hline
\end{tabularx}
\end{table}


\vspace{-0.3cm}
\subsection{Foot Position and Required Rotation}

\begin{figure}[tbhp]
    \centering
    \includegraphics[width=1.0\columnwidth]{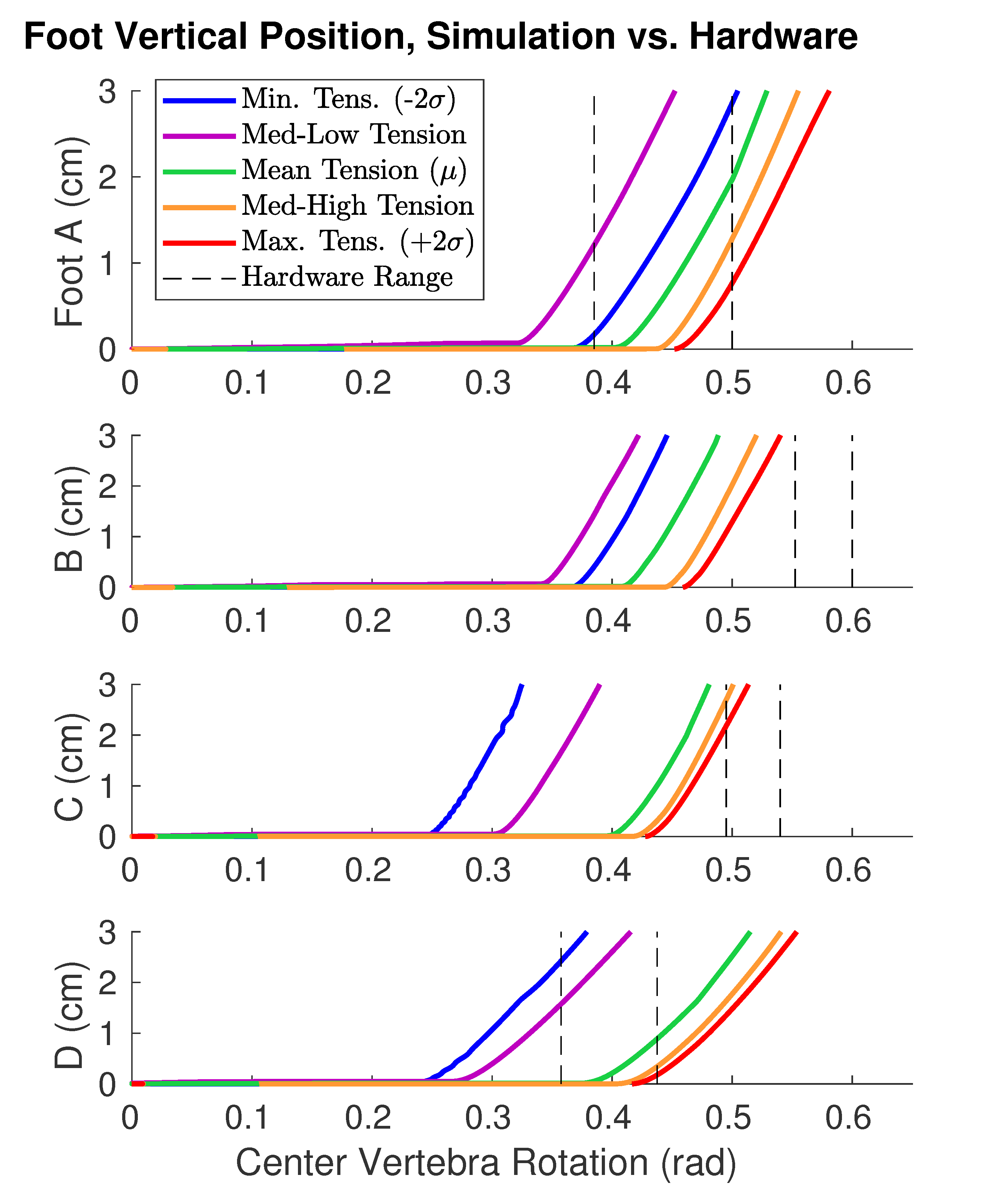}
    \vspace{-0.4cm}
    \caption{Simulations and hardware results of foot-lifting tests for Laika's spine. Black dashed lines represent the range of lift-off points in hardware. Colored curves represent vertical position of each foot (A, B, C, D) at varying levels of lattice tension in simulation. The highest-tension simulation result (red) matches hardware most closely, and represents a calibration of the simulation for future work.}
    \label{fig:footpos_all20}
    \vspace{-0.2cm}
\end{figure}

The results of five hardware tests per foot are plotted against the simulation results in Fig. \ref{fig:footpos_all20}.
The center-vertebra rotations at foot lift-off, observed in hardware, are plotted as black vertical lines representing the minimum and maximum datapoints.
Colored curves are simulation data at the different lattice tension levels from Table \ref{tab:simulation_tests}.

The rotations for foot lift-off are summarized in Table \ref{tab:liftoff_rotations}.
The hardware minimum and maximum correspond to the black dashed lines in Fig. \ref{fig:footpos_all20}.
The simulation data listed are the points where each curve leaves the $y$-axis in Fig. \ref{fig:footpos_all20}, with the variation arising from lattice tension.

\begin{table}[th]
\centering
\caption{Range of center vertebra rotations that produced foot lift-off, in simulation and hardware. All angles in radians (abs. val.)}
\label{tab:liftoff_rotations}
\begin{tabularx}{1\columnwidth}{| l | X  X  X  X |}
    \hline
    Foot & Simul., Min & Simul., Max & HW, Min & HW, Max \\
    \hline
    A & 0.33 & 0.47 & 0.44 & 0.50 \\
    B & 0.35 & 0.47 & 0.57 & 0.60 \\
    C & 0.25 & 0.44 & 0.51 & 0.54 \\
    D & 0.25 & 0.43 & 0.41 & 0.43 \\
    \hline
\end{tabularx}
\vspace{-0.2cm}
\end{table}


Across all tests, the maximum-tension lattice (red lines in Fig. \ref{fig:footpos_all20}) was most representative of hardware, falling within the range for feet A and D and closest to the range for feet B and C.
The simulation results for feet B and C fall close to the hardware range, but not within.
Such results can be expected with the small amount of testing that was performed in hardware, and with the variation in the test setup.
\section{DISCUSSION AND FUTURE WORK}


Simulation data produced a calibration (highest lattice tension) that can reasonably be used for future work in developing balancing motions and gait cycles of the robot.
The error in the B/C foot simulations, which did not strictly lie within the hardware range, can be attributable to the simplifications made in the simulation model.
These simplifications include a combination of frictional effects at the robot's feet, the variation due to manual tensioning of the hardware robot's cables, and the simplified geometry in the simulation.

\subsection{Differences Between Foot Lift-Off Behaviors}

In addition to the differences observed between simulation and hardware for feet B and C, there were also differences between each foot with respect to lift-off angle as well as height after lift-off.
These differences are expected, due to the spine's geometry.

Feet C and D, the back feet, lifted with less rotation than their front-opposite counterparts (C versus B, and D versus A.)
This anterior-posterior difference is attributed to Laika's asymmetry in that direction, with more weight (due to the spine) at the robot's shoulders (Fig. \ref{fig:assembly}.)

Feet B and C, which lifted with clockwise rotation, raised more rapidly after the initial lift-off.
Such a difference is expected due to the geometry of the robot's saddle cables. 
These cables do not lie completely in the transverse plane of the spine: they pull the vertebrae forward and backward as well (Fig. \ref{fig:spine_topology}.)
Thus, when the center vertebra rotates, it also creates a small amount of additional bending in both the horizontal (coronal) and front-back vertical (sagittal) plane, as its saddle cables adjust.
The clockwise/counterclockwise difference is attributed to center-of-mass shifts in the transverse (left-right vertical) plane from this extra bending.

\subsection{Future Work}

These preliminary results show a proof-of-concept hardware prototype of Laika's spine, with simulation results corresponding to hardware motions.
These results show that the simulation is suitable for developing balancing and walking gaits for future versions of the robot, when actuated legs are included.
Future robot designs will require changes in cable tensions, lateral bending amounts, and terrain geometry, all of which will be considered when building walking models of Laika with its spine.






\section*{ACKNOWLEDGMENT}


This work would not have been possible without the help of the many members of the Berkeley Emergent Space Tensegrities Lab and the Dynamic Tensegrity Robotics Lab at NASA Ames Research Center's Intelligent Robotics Group.

This research was supported by NASA Prime Contract no. NAS2-03144, NASA Early Stage Innovation Grant NNX15AD74G, NASA Space Technology Research Fellowship no. NNX15AQ55H, and NSF Graduate Research Fellowship no. DGE1106400.

\bibliographystyle{IEEEtran}
\bibliography{bibliographies/library}


\balance

\end{document}